\documentclass{llncs}

\usepackage{mathptmx}       
\usepackage{helvet}         
\usepackage{courier}        
\usepackage{type1cm}        
\usepackage{amsmath}
\usepackage{makeidx}         
\usepackage{graphicx}        
\usepackage{subcaption}          
\usepackage{multicol}        
\usepackage[bottom]{footmisc}
\usepackage{amsmath}
\usepackage{amsfonts}
\usepackage{amssymb}
\usepackage{epstopdf}
\usepackage{float}
\usepackage{hyperref}
\usepackage[flushleft]{threeparttable}

\makeindex             

\begin{document}
\title{AN OVERVIEW OF CAUSAL INFERENCE USING KERNEL EMBEDDINGS}
\author{Dino Sejdinovic}
\institute{ 
              University of Adelaide, Australia \\
             \email{dino.sejdinovic@adelaide.edu.au}                  
            }            
\maketitle
\begin{abstract}
Kernel embeddings have emerged as a powerful tool for representing probability measures in a variety of statistical inference problems. By mapping probability measures into a reproducing kernel Hilbert space (RKHS), kernel embeddings enable flexible representations of complex relationships between variables. They serve as a mechanism for efficiently transferring the representation of a distribution downstream to other tasks, such as hypothesis testing or causal effect estimation. In the context of causal inference, the main challenges include identifying causal associations and estimating the average treatment effect from observational data, where confounding variables may obscure direct cause-and-effect relationships. Kernel embeddings provide a robust nonparametric framework for addressing these challenges. They allow for the representations of distributions of observational data and their seamless transformation into representations of interventional distributions to estimate relevant causal quantities. We overview recent research that leverages the expressiveness of kernel embeddings in tandem with causal inference.

\end{abstract}
 
\section{Introduction}
\label{sec:1}

Kernel embeddings \cite{muandetKernelMeanEmbedding2017a} have gained prominence as a versatile tool for addressing complex challenges in modern statistical inference. At their core, kernel embeddings enable the representation of probability measures in a reproducing kernel Hilbert space (RKHS) \cite{aronszajnTheoryReproducingKernels1950}, and facilitate nonparametric approaches to a wide array of statistical problems. This framework has emerged as particularly potent when dealing with scenarios that involve intricate dependencies between variables, where traditional parametric methods often falter. The flexibility of kernel embeddings opens up new possibilities for manipulating probability measures in order to address downstream tasks.

One area where kernel embeddings have shown their potential is in causal inference. Causal inference is challenging, especially in the presence of unobserved confounding variables that obscure the direct relationships between causes and effects. Standard parametric methods often rely on strong assumptions about the data generating processes -- assumptions that, when violated, can lead to erroneous conclusions. Kernel embeddings, on the other hand, offer a more flexible, nonparametric approach that sidesteps many of these restrictive assumptions, making them a valuable tool for tackling causality in complex, real-world datasets.

In particular, kernel embeddings provide a way to encode observational distributions in a manner that allows for seamless transformations into interventional distributions, which are crucial for causal reasoning. As a result, we can derive estimators of key causal quantities such as the average treatment effect (ATE) or distributional treatment effect (DTE) from observational data or test for the presence of causal associations, without the need for parametric specifications.

In this paper, we provide an overview of the diverse and growing body of research that employs kernel embeddings for causal inference. Our aim is to provide a coherent and systematic introduction to the foundational concepts. We put emphasis on expressing distributional treatment effects using conditional and deconditional mean operators, which can be used as a unified framework for building nonparametric estimators of ATE and DTE.

\section{Background on Kernel Embeddings}
\subsection{RKHS and feature maps}
A reproducing kernel Hilbert space (RKHS) \cite{aronszajnTheoryReproducingKernels1950,berlinetReproducingKernelHilbert2004} is a Hilbert space $\mathcal H$ of functions $f:\mathcal X\to \mathbb R$ where evaluation $f\mapsto f(x)$ is a continuous linear functional for every point $x\in\mathcal X$. The domain $\mathcal X$ is a generic non-empty set. By Riesz representation theorem, an RKHS has a unique \emph{reproducing kernel}, a function $k : \mathcal X \times \mathcal X \to \mathbb R$, such that $\forall x\in\mathcal{X},\;\;k(\cdot,x)\in\mathcal{H}$, and $\forall x\in\mathcal{X},\,\forall f\in\mathcal{H},\;\;\left\langle f,k(\cdot,x)\right\rangle_{\mathcal{H}}=f(x)$ (\emph{the reproducing property}). Conversely, any positive definite function $k$ is a reproducing kernel for a unique RKHS $\mathcal H_k$ -- this result is known as the Moore-Aronszajn theorem \cite{berlinetReproducingKernelHilbert2004}. The formalism of kernel methods in machine learning \cite{hofmannKernelMethodsMachine2008} relies on implicit representations
of individual data points, via the canonical feature map $\phi\colon x\mapsto k\left(\cdot,x\right)$,
such that every data point is represented as a point in the RKHS $\mathcal{H}_k$. Linear algorithms within $\mathcal{H}_k$ as a feature space result in nonlinear solutions in the original domain, transforming complex relationships into tractable problems without explicit mapping into $\mathcal{H}_k$.

\subsection{Embeddings and MMD}
Kernel embeddings  \cite{muandetKernelMeanEmbedding2017a} represent probability measures in the
RKHSs by considering the expectation of the canonical feature map. Kernel embedding for a random variable $X$ with law $P$ and for a given kernel $k$ is

\[
\mu_{k,P}=\mathbb{E}_{X\sim P}k\left(\cdot,X\right)\in\mathcal{H}_k.
\]
We will suppress the dependence on $k$ in notation, and often also index embedding with the corresponding random variable $X$, i.e. as $\mu_X$, when there is no ambiguity.

Kernel embedding is a potentially infinite-dimensional representation of the probability measure $P$, which is
akin to a characteristic function. Kernel embedding is a Riesz representer of the expectation functional over RKHS, i.e.

\[
\left\langle f,\mu_{P}\right\rangle _{\mathcal{H}_k}=\mathbb{\mathbb{E}}_{X\sim P}f(X),\quad\forall f\in\mathcal{H}_k
\]
and it exists whenever $f\mapsto\mathbb{\mathbb{E}}_{X\sim P}f(X)$ is a bounded
functional. Note that this is always true if the kernel function itself
is bounded, i.e. $k(x,y)\leq M<\infty$ $\forall x,y$, by
Cauchy-Schwarz,
\[
\mathbb{\mathbb{E}}_{X\sim P}f(X)=\mathbb{\mathbb{E}}_{X\sim P}\left\langle f,k\left(\cdot,X\right)\right\rangle_{\mathcal H_k} \leq\left\Vert f\right\Vert_{\mathcal{H}_k}\mathbb{E}_{X\sim P}\left\Vert k\left(\cdot,X\right)\right\Vert_{\mathcal{H}_k}\leq\sqrt{M}\left\Vert f\right\Vert_{\mathcal{H}_k}.
\]

Kernel embeddings impose a simple Hilbert space structure on probability measures. In particular, inner products between kernel embeddings
can be computed as

\[
\left\langle \mu_P,\mu_Q\right\rangle _{\mathcal{H}_{k}}=\mathbb{E}_{X\sim P}\mathbb{E}_{Y\sim Q}k(X,Y),
\]
and we can easily estimate the (squared) distances between probability
measures induced by this RKHS representation since they correspond
to simple expectations. Such distances are called \emph{Maximum Mean
Discrepancy (MMD)}:

\begin{eqnarray}
\text{MMD}_{k}^{2}\left(P,Q\right) & = & \left\Vert \mu_P-\mu_Q\right\Vert _{\mathcal{H}_{k}}^{2}\label{eq: popMMD}\\
 & = & \mathbb{E}_{X,X'\sim P}k(X,X')+\mathbb{E}_{Y,Y'\sim Q}k(Y,Y')-2\mathbb{E}_{X\sim P,Y\sim Q}k(X,Y),\nonumber 
\end{eqnarray}
where $X$ and $X'$ denote independent copies of random variables
with law $P$, and similarly for $Y$ and $Y'$. For a large class of kernels, including Gaussian and Mat\'ern families of kernels, MMD is a proper metric on probability measures,
in the sense that $\text{MMD}_{k}\left(P,Q\right)=0$ implies $P=Q.$ Such
kernels are called \emph{characteristic} \cite{sriperumbudurUniversalityCharacteristicKernels2011}. MMD is a popular probability
metric, widely used for nonparametric hypothesis testing \cite{grettonKernelTwoSampleTest2012} and various machine learning applications.

\subsection{Product kernels and HSIC}
Given kernels $k_{\mathcal{X}}$
and $k_{\mathcal{Y}}$ on the respective domains $\mathcal{X}$ and
$\mathcal{Y}$, we can define a kernel on the product domain $\mathcal{X}\times\mathcal{Y}$, $k_{\mathcal X\times \mathcal Y}=k_{\mathcal{X}}\otimes k_{\mathcal{Y}}$,
by
\[
k_{\mathcal{X}\times\mathcal{Y}}\left(\left(x,y\right),\left(x',y'\right)\right)=k_{\mathcal{X}}(x,x')k_{\mathcal{Y}}(y,y').
\]
The tensor notation signifies that
the canonical feature map of $k_{\mathcal{X}\times\mathcal{Y}}$ is $(x,y)\mapsto k_{\mathcal{X}}(\cdot,x)\otimes k_{\mathcal{Y}}(\cdot,y)$.
The RKHS of the kernel $k_{\mathcal{X}}\otimes k_{\mathcal{Y}}$
is in fact isometric to $\mathcal{H}_{k_{\mathcal{X}}}\otimes\mathcal{H}_{k_{\mathcal{Y}}}$,
which can be viewed as the space of Hilbert-Schmidt operators between
$\mathcal{H}_{k_{\mathcal{Y}}}$ and $\mathcal{H}_{k_{\mathcal{X}}}$.

\emph{Hilbert-Schmidt Independence Criterion (HSIC)}
$\text{HSIC}_{k_{\mathcal{X}},k_{\mathcal{Y}}}(X,Y)$ measures dependence between random variables $X$ and $Y$ using
the MMD between the joint measure $P_{XY}$ and the product
of marginals $P_{X}P_{Y}$, computed with the product kernel $k_{\mathcal{X}}\otimes k_{\mathcal{Y}}$,
i.e., 
\begin{eqnarray*}
\text{HSIC}_{k_{\mathcal{X}},k_{\mathcal{Y}}}(X,Y) & = & \left\Vert \mu_{P_{XY}}-\mu_{P_{X}P_{Y}}\right\Vert _{\mathcal{H}_{k_{\mathcal{X}}\otimes k_{\mathcal{Y}}}}\\
 & = & \left\Vert \mathbb{E}_{XY}[k_{\mathcal{X}}(.,X)\otimes k_{\mathcal{Y}}(.,Y)]-\mathbb{E}_{X}k_{\mathcal{X}}(.,X)\otimes\mathbb{E}_{Y}k_{\mathcal{Y}}(.,Y)\right\Vert _{\mathcal{H}_{k_{\mathcal{X}}\otimes k_{\mathcal{Y}}}}.
\end{eqnarray*}
A sufficient condition for HSIC to be well defined is that both kernels
$k_{\mathcal{X}}$ and $k_{\mathcal{Y}}$ are bounded. The name
HSIC comes from identifying $\mu_{P_{XY}}-\mu_{P_{X}P_{Y}}\in \mathcal{H}_{k_{\mathcal{X}}\otimes k_{\mathcal{Y}}}$ 
with the \emph{cross-covariance operator} $C_{XY}:\mathcal{H}_{k_{\mathcal{Y}}}\to\mathcal{H}_{k_{\mathcal{X}}}$
for which 
\[
\langle f,C_{XY}g\rangle_{\mathcal{H}_{k_{\mathcal{X}}}}=\text{Cov}\left[f(X)g(Y)\right],\quad\forall f\in\mathcal{H}_{k_{\mathcal{X}}},g\in\mathcal{H}_{k_{\mathcal{Y}}},
\]
whereby HSIC is the Hilbert-Schmidt norm of $C_{XY}$. Note that this property is analogous to the finite-dimensional one, $a^{\ensuremath{\top}}C_{XY}b=\text{Cov}\left[a^{\top}X,b^{\top}Y\right]$,
where $X$ and $Y$ are random vectors and $C_{XY}$ is their cross-covariance
matrix. HSIC is closely related to the notion of distance correlation \cite{szekelyMeasuringTestingDependence2007,sejdinovicEquivalenceDistancebasedRKHSbased2013}.

\subsection{Conditional Mean Embeddings and Operators}
Given random variables $X, Y$ with joint distribution $P_{XY}$, the conditional mean embedding (CME) \cite{songKernelEmbeddingsConditional2013a} with respect to the conditional distribution $P(Y|X=x)$, is defined as
\begin{align}
    \mu_{Y|X=x}:=\mathbb E_{Y|X=x}[k_{\mathcal Y}(\cdot,Y)]\in \mathcal H_{k_\mathcal Y}.
\end{align}
CME can be viewed as a solution to a ridge regression between feature spaces \cite{grunewalderConditionalMeanEmbeddings2012a}. Given a dataset $\mathcal D=\{(X_i,Y_i)\}_{i=1}^n \sim P_{XY}$, a standard estimator of CME is
\[
\hat{\mu}_{Y|X=x}=\sum_{i=1}^n \beta_{i}(x)k_{\mathcal Y}(\cdot,Y_{i}),
\]
with the vector of coefficients $\beta(x)$ computed similarly to standard ridge regression formula as
\[
\beta(x)=(K_{\mathcal X}+n\lambda I_{n})^{-1}k_{\mathcal X}(\mathbf{X},x).
\]
Here, $K_{\mathcal X}$ is an $n\times n$ matrix with $[K_{\mathcal X}]_{ij}=k_{\mathcal X}(X_i,X_j)$, and $k_{\mathcal X}(\mathbf{X},x)$ is an $n\times 1$ vector with $[k_{\mathcal X}(\mathbf{X},x)]_i=k_{\mathcal X}(X_i,x)$.

The \textit{conditional mean operator} (CMO) $C_{Y|X}:\mathcal H_{k_\mathcal X}\to\mathcal H_{k_\mathcal Y}$ models how CMEs vary as a function of the conditioning variable $x$. CMO satisfies
\[
C_{Y|X}k(\cdot,x)=\mu_{Y|X=x},
\]
or, equivalently, in terms of its adjoint operator, $C_{Y|X}^*f=\mathbb E[f(Y)|X=\cdot]$. While the often used expression $C_{Y|X}=C_{YX}C_{XX}^{-1}$ develops intuition that CMO can be linked to the (uncentred) cross-covariance operators, the inverse on the right hand side is not well defined in infinite-dimensional spaces. An alternative measure-theoretic approach for conditional mean embeddings was developed in \cite{parkMeasureTheoreticApproachKernel2020a}. The conditions under which CMO in infinite-dimensional spaces is well defined, as well as the convergence properties of the corresponding estimators, are thoroughly studied elsewhere (e.g. \cite{liOptimalRatesRegularized2022}) and we do not focus on these aspects here, in order to maintain clarity of exposition on how the framework of CMOs (assuming they are well defined) can be applied to causal inference. The standard CMO estimator is given by
\[
\hat{C}_{Y|X}=\Psi_{\mathbf{Y}}\left(K_{\mathcal X}+n\lambda I_{n}\right)^{-1}\Phi_{\mathbf{X}}^*,
\]
where  $\Phi_{\mathbf{X}}$ is understood as the operator
\[
 \Phi_{\mathbf{X}}:\mathbb{R}^n\to \mathcal H_{k_\mathcal X},\quad \Phi_{\mathbf{X}}\beta=\sum_{i=1}^n \beta_{i}k_{\mathcal X}(\cdot,X_{i}),
 \]
 and similarly for $\Psi_{\mathbf{Y}}$.

The regression view of conditional mean embeddings \cite{grunewalderConditionalMeanEmbeddings2012a} can be further exploited to construct more flexible representations via deep learning which also circumvent the quadratic computational cost of kernel ridge regression \cite{ShiFukSej2024}. More broadly, combining kernel embeddings with deep learning is a promising and active area of research \cite{lawHyperparameterLearningDistributional2019,xuLearningDeepFeatures2021,nguyen-tangDistributionalReinforcementLearning2021,xuDeepProxyCausal2021,xuNeuralMeanEmbedding2022}.  

\subsection{Bayesian Kernel Embeddings}

Quantifying epistemic uncertainty in the representation of the unknown probability measure is essential in a wide range of applications, where uncertainty needs to propagate into a downstream task, such as active learning or Bayesian optimization \cite{garnettBayesianOptimization2023}. Given that kernel embedding is a function in a Hilbert space, a natural model to consider is a Gaussian process \cite{rasmussenGaussianProcessesMachine2005}, however some technical challenges related to formulating Gaussian process priors on reproducing kernel Hilbert spaces, so called Driscoll's 0/1 laws \cite{driscollReproducingKernelHilbert1973} need to be surmounted to construct the appropriate Bayesian framework. Bayesian embeddings were first studied in \cite{flaxmanBayesianLearningKernel2016}, and refined for the problem of Bayesian two-sample testing in \cite{zhangBayesianKernelTwoSample2022}. Furthermore, the formulation was extended to conditional mean embeddings in \cite{chauBayesIMPUncertaintyQuantification2021} specifically for the purposes of causal data fusion, which we will discuss in Section \ref{sec:fusion}.

\subsection{Deconditional Mean Embeddings}
\label{sec:deconditioning}
Deconditional mean embeddings were introduced by \cite{hsuBayesianDeconditionalKernel2019} as a counterpart to CMEs. While CME $\mu_{Y|X=x}$ allows us to take the conditional expectation for $f\in \mathcal{H}_{k_\mathcal{Y}}$ as an inner product
\[
\mathbb{E}\left[f(Y)|X=x\right]=\langle f,\mu_{Y|X=x} \rangle_{\mathcal{H}_{k_\mathcal{Y}}},
\]
i.e. CME represents the mapping $f\mapsto \mathbb{E}\left[f(Y)|X=\cdot\right]$, a deconditional mean embedding (DME) denoted $\eta_{Y=y|X}$, represents the inverse mapping, i.e. deconditioning $\mathbb{E}\left[f(Y)|X=\cdot\right]\mapsto f$. Thus, DME reconstructs the initial function of which the conditional expectation was taken through the inner product, as
\[
\langle \mathbb{E}\left[f(Y)|X=\cdot\right],\eta_{Y=y|X} \rangle_{\mathcal{H}_{k_\mathcal{X}}}=f(y). 
\]
The associated operator is deconditional mean operator (DMO), $D_{Y|X}: \mathcal{H}_{k_\mathcal{Y}}\to \mathcal{H}_{k_\mathcal{X}}$ such that $D_{Y|X}k_{\mathcal{Y}}(\cdot,y)=\eta_{Y=y|X}$. Similarly to CMOs, DMOs require certain conditions to be well defined in infinite-dimensional cases, as discussed in \cite{hsuBayesianDeconditionalKernel2019}. Estimators for DMOs are given in \cite{hsuBayesianDeconditionalKernel2019} and also can be linked to a formulation of deconditioning using Gaussian processes developed in \cite{chauDeconditionalDownscalingGaussian2021a}, which was motivated by statistical downscaling problems. Convergence properties of DMO estimators are also presented in \cite{chauDeconditionalDownscalingGaussian2021a}.

\section{Background on Causal Inference}
There are two main approaches to causal inference: the framework  of potential outcomes / counterfactuals, also known as Neyman-Rubin causal models, pioneered by Donald Rubin \cite{rubinEstimatingCausalEffects1974a}; and graphical causal inference, pioneered by Judea Pearl, which relies on concepts like Directed Acyclic Graphs (DAGs) and do-Calculus \cite{pearlCausalityModelsReasoning2000}. Single World Intervention Graphs (SWIGs) \cite{richardsonSingleWorldIntervention2013} are a way to unify counterfactual and graphical approaches. In this note, we mainly focus on graphical causal inference but will also briefly mention some connections to the potential outcomes framework.

The literature on causal inference is very broad and often uses a myriad of different conventions and notations. Here we aim to give a unified presentation using consistent notation on a selection of topics. We denote the outcome of interest by $Y$, the treatment variable by $T$, the potential outcomes under treatment $t$ by $Y(t)$, the observed confounders which form an adjustment set by $X$, unobserved confounders by $\xi$, mediators by $S$, additional covariates / effect modifiers by $V$, instruments / treatment proxies by $Z$, and the outcome proxies by $U$. The presentation will be general in the sense that each of $S$, $T$, $U$, $V$, $X$, $Y$, $Z$ can be discrete or continuous, can correspond to a random vector, a set of random variables, or a random quantity taking values in a generic domain where Borel probability measures can be defined. We refrain from using the symbol $W$ for random variables, as we reserve it for weights in the approaches that rely on weighting. The domain for $S$ will be denoted $\mathcal S$, kernel on $\mathcal S$ will be denoted $k_\mathcal S$ and corresponding RKHS $\mathcal H_{\mathcal S}$, and similarly for $T$, $U$, $V$, $X$, $Y$, $Z$.  We will also use lower case $p$ to denote either the probability mass function or the probability density function (assuming for simplicity that all probability measures considered have densities). 
\subsection{Causal graphs and do-Operator}
\begin{figure}[ht]
    \centering
    \begin{subfigure}[b]{0.45\textwidth}
        \centering
        \includegraphics[width=\textwidth]{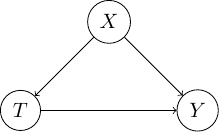}
        \caption{Observational distribution $p$}
        \label{fig:backdoor-a}
    \end{subfigure}
    \hfill
    \begin{subfigure}[b]{0.45\textwidth}
        \centering
        \includegraphics[width=\textwidth]{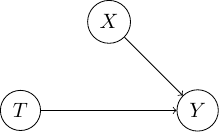}
        \caption{Interventional distribution $p^\star$}
        \label{fig:backdoor-b}
    \end{subfigure}
    
    \caption{Backdoor adjustment}
    \label{fig:backdoor}
\end{figure}

Graphical causal inference uses Directed Acyclic Graphs (DAGs) to capture a particular factorization of the joint distribution of random variables. DAGs are a broader concept and need not be interpreted causally, so we use them in conjunction with the so called \textit{(strict) causal edges assumption}: in a causal DAG, \textit{every parent is a direct cause of all its children}. An example of a DAG is given in Fig. \ref{fig:backdoor-a}, where we represent that the treatment $T$ and the confounder $X$ are both direct causes of the outcome of interest $Y$, and that the confounder $X$ also causally affects the treatment $T$. Interventions are denoted using the \textit{do-operator}, $do(T=t)$, indicating that we are looking into the distribution of the outcomes $Y$, where everyone in the population is given the treatment $t$. This is called the \textit{interventional distribution} $p(Y|do(T=t))$. In contrast, conditioning on $T=t$ means that we are looking at the outcomes for the subset of the population who have received the treatment $t$ in the observational data so $p(Y|do(T=t))\neq p(Y|T=t)$. Intervention is a form of \textit{graph manipulation}. When we make an intervention on the treatment variable set $T$ (which may correspond to multiple nodes in a DAG), we construct a new DAG where we have removed all of the incoming edges to nodes in the set $T$ (like in Fig. \ref{fig:backdoor-b}). This construction formulates a new joint distribution of variables of interest, representing the relationships following an intervention, in contrast to the original DAG which corresponds to the observational distribution from which data is collected. For the setting given in Fig. \ref{fig:backdoor-a}, the joint distribution given by DAG is given by
\[
p(X,T,Y)=p(X)p(T|X)p(Y|X,T).
\]

Following the intervention $do(T=t)$, we obtain the joint interventional distribution over remaining variables as
\[
p(X,Y|do(T=t))=p(X)\mathbf{1}\{T=t\}p(Y|X,T=t).
\]
Integrating $X$ out gives the well known \emph{backdoor formula}:
\[
p(Y|do(T=t))=\int p(Y|X,T=t)p(X)dX.
\]
We can also view interventions as ``soft'' and consider the joint distribution where $T$ has some arbitrary marginal $p^{\star}(T)$, giving the new joint for the world of interventions in Fig. \ref{fig:backdoor-b}:
\[
p^{\star}(X,T,Y)=p(X)p^{\star}(T)p(Y|X,T).
\]
The conditional $p^\star(Y|T=t)$ under this new distribution $p^\star$ is precisely the interventional distribution $p(Y|do(T=t))$ we wish to estimate.

The framework for expressing probabilities involving do-operator using ordinary conditional probabilities is known as \textit{do-Calculus}.  When focusing on \textit{nonparametric identification}, i.e. the setting where only assumptions made are those encoded in DAGs, remarkably, there exist three simple rules of do-calculus which are sufficient to identify all identifiable causal estimands. The proofs for this so called completeness of do-Calculus are also constructive, which means that they admit algorithms that identify any causal estimand in polynomial time \cite{shpitserIdentificationJointInterventional2006,huangPearlsCalculusIntervention2006}. 

\subsection{Potential Outcomes}
Whilst also general, the potential outcomes framework is often presented for the case of a binary treatment, $T\in\{0,1\}$. For each unit, one defines two random variables, called potential outcomes, under treatment ($Y(1)$) and under control ($Y(0)$). The \textit{treatment effect} is the difference between these two potential outcomes, $Y(1)-Y(0)$. Since it is possible to observe only one potential outcome per unit, the treatment effect is not directly observable. This is referred to as \textit{the fundamental problem of causal inference}. The observed outcome $Y$ is defined as

$$
Y=\mathbf{1}\{T=0\}Y(0)+\mathbf{1}\{T=1\}Y(1).
$$
Importantly, the treatment variable $T$ generally depends on $Y(0)$, $Y(1)$, and on (measured or unmeasured) covariates of the unit. More general treatment $t\in\mathcal{T}$ for some domain $\mathcal T$ can be linked to do-Calculus by identifying the potential outcome $Y(t)$ with $Y|do(T=t)$.

\subsection{Average Treatment Effect}
An important quantity of interest in causal inference is the average treatment effect (ATE), given by
$$
f(t)=\mathbb{E}\left[Y|do(T=t)\right]=\mathbb{E}\left[Y(t)\right].
$$
We are also sometimes interested in the conditional average treatment effect (CATE)
$$
f(t,v)=\mathbb{E}\left[Y|do(T=t),V=v\right]=\mathbb{E}\left[Y(t)|V=v\right],
$$
where $V$ are covariates / effect modifiers which identify a particular subpopulation of interest. CATE is also sometimes referred to as the heterogeneous response curve \cite{singhKernelMethodsCausal2024}. We note that sometimes in the literature there is a conflation between effect modifiers denoted by $V$ and the confounders that correspond to a sufficient adjustment set which we shall denote by $X$. 
When treatment is binary, we usually write (C)ATE in terms of the difference between the potential outcomes, i.e.
$$
f(1,v)-f(0,v)=\mathbb{E}\left[Y({1})-Y({0})|V=v\right].
$$
The framework of do-Calculus allows appropriately adjusting standard regression to estimate ATE and CATE in a number of causal DAGs. Another important building block for estimating causal effects are weighting-based estimators \cite{jungEstimatingCausalEffects2020}. 
\subsection{Distributional Treatment Effects}
Instead of estimating expectations of the outcome $Y$ under different interventions, we may be interested in a richer representation of the distribution of the outcome $Y$, e.g. using quantiles or probability densities, or indeed, using the corresponding kernel embeddings. In particular, we can define the distributional treatment effect (DTE) which we will also refer to as the interventional mean embedding (IME) as
\begin{align}
    \mu_{Y|do(T=t)}:=\mathbb E_{Y|do(T=t)}[k_{\mathcal Y}(\cdot,Y)]\in \mathcal H_{\mathcal Y}.
\end{align}
Analogously, the conditional version, CDTE (CIME), is given by
\begin{align}
    \mu_{Y|do(T=t),V=v}:=\mathbb E_{Y|do(T=t),V=v}[k_{\mathcal Y}(\cdot,Y)]\in \mathcal H_{\mathcal Y},
\end{align}
where $V$ are again the effect modifiers. Estimating IMEs and their distances via MMD was studied by \cite{muandetCounterfactualMeanEmbeddings2021,FawHuEvaSej2024} for the potential outcomes under binary treatment. When considering the continuous treatment variable / which can be thought of as a dose, various quantities of interest can be approached using the kernel embeddings. A comprehensive study was presented in \cite{singhKernelMethodsCausal2024}. In this note, we express the IMEs in the operator form, using CMOs and DMOs, which unifies and significantly simplifies the construction of estimators in various settings. For this purpose, we will also define an interventional mean operator (IMO), $C_{Y|do(T)}:\mathcal{H}_{\mathcal{T}}\to \mathcal{H}_{\mathcal{Y}}$ for which
$C_{Y|do(T)}^{*}g=\mathbb{E}\left[g(Y)|do(T=\cdot)\right],$
and similarly if conditioning on effect modifiers $V$.

We first consider the most straightforward setting, \textit{backdoor adjustment} in Fig.\ref{fig:backdoor}, where we assume that all confounders are observed, followed by the setting which allows for ATE estimation from unmatched datasets (Section \ref{sec:fusion}). We then turn our attention to unobserved confounding, in three distinct settings, frontdoor in Section \ref{sec:frontdoor}, instruments in Section \ref{sec:instruments}, and proxies in Section \ref{sec:proxies}. In each of the settings we describe, we will derive an expression of DTE using CMOs and DMOs which allows straightforward construction of corresponding estimators. The summary of these results is given in Table \ref{tab:summary}. The estimators presented for the basic settings of backdoor and frontdoor adjustment have been considered in several works and their convergence properties were studied in \cite{singhKernelMethodsCausal2024}, who also give an example of their applications to empirical economics. The setting of causal data fusion follows closely \cite{chauBayesIMPUncertaintyQuantification2021}, extending it from ATE to DTE. A kernel version of instrumental variable regression was considered in \cite{singhKernelInstrumentalVariable2019} and its deep feature version in \cite{xuLearningDeepFeatures2021}, and similarly for proxies \cite{mastouriProximalCausalLearning2021,xuDeepProxyCausal2021}.  

\begin{table}[]
\begin{center}
\begin{tabular}{|l|l|l|}
\hline
\textbf{Setting}     & \textbf{IME} $\mu_{Y|do(T=t)}$  \\ \hline
\textit{backdoor}, Fig. \ref{fig:backdoor}       &  $C_{Y|T,X}(k_{\mathcal{T}}(\cdot,t)\otimes\mu_{X})$   \\ \hline
\textit{fusion}, Fig. \ref{fig:bayesimp}      & $C_{Y|S}C_{S|T,X}(k_{\mathcal{T}}(\cdot,t)\otimes\mu_{X})$       \\ \hline
\textit{frontdoor}, Fig. \ref{fig:frontdoor}       & $C_{Y|S,T}((C_{S|T}k_{\mathcal{T}}(\cdot,t))\otimes\mu_{T})$        \\ \hline
\textit{instruments}, Fig. \ref{fig:instrument}       &  $C_{Y|Z}D_{T|Z}k_{\mathcal T}(\cdot,t)$   \\ \hline
\textit{proxies}, Fig. \ref{fig:proxies}       & $C_{Y|T,Z}D_{(T,U)|(T,Z)}(k_{\mathcal T}(\cdot,t)\otimes \mu_{U})$        \\ \hline
\end{tabular}
\end{center}
\caption{Summary of interventional mean embeddings (IME) in different settings}
\label{tab:summary}
\end{table}

\section{Backdoor adjustment}
\label{sec:backdoor}

\subsection{Backdoor criterion and DTE estimation}
Backdoor adjustment is applied in the setting where confounder is fully observed, namely there exists a set of variables which satisfies a particular graphical criterion. The set of variables $X$ is said to be a \textit{sufficient backdoor adjustment set} if it satisfies the following \textit{backdoor criterion} relative to the pair $(T,Y)$ of treatment $T$ and outcome $Y$: (a) $X$ blocks all backdoor paths from $T$ to $Y$, (b) $X$ does not contain any descendants of $T$. A sufficient backdoor adjustment set is \textit{not unique} and one can consider optimality criteria for such sets \cite{rungeNecessarySufficientGraphical2021}. The backdoor criterion is stated in the potential outcome language as the assumption typically referred to as \textit{unconfoundedness}: $(Y(0),Y(1))\perp\!\!\!\!\perp T|X$. Given a sufficient backdoor adjustment set $X$, we can relate the interventional distribution $Y|do(T=t)$ to the conditional distribution $Y|T,X$ as
$$
p(Y|do(T=t))=\int p(Y|T=t,X)p(X)dX.
$$
Of course, the conditional distribution $Y|T,X$ can be estimated from observational data consisting of the matched triplet of confounder, treatment, and outcome $\{(X_i,T_i,Y_i)\}_{i=1}^n$.

Applying the backdoor criterion to kernel embeddings, we get
\begin{align*}
\mu_{Y|do(T=t)}&=\int \mu_{Y|T=t,X=x}p(x)dx\\
&=\int C_{Y|T,X}(k_{\mathcal{T}}(\cdot,t)\otimes k_{\mathcal{X}}(\cdot,x))p(x)dx\\
&=C_{Y|T,X}(k_{\mathcal{T}}(\cdot,t)\otimes\mu_{X}).
\end{align*}

Now, to construct an estimator of DTE, we can simply plug in the empirical estimators of $C_{Y|T,X}$ and $\mu_X$,
\[
\hat C_{Y|T,X} =  \Psi_{\mathbf{Y}}\left(K_{\mathcal X}\circ K_{\mathcal T} +n\lambda I_{n}\right)^{-1}\Phi_{\mathbf{X,T}}^*,\qquad \hat\mu_X=\frac{1}{n}\sum_{i=1}^n k_{\mathcal X}(\cdot,X_i).
\]
Putting them together, we obtain
\[
\hat{\mu}_{Y|do(T=t)}=\sum_{j=1}^{n} \gamma_{i}(t)k_{\mathcal{Y}}(\cdot,Y_{i}),
\]
i.e. IME is, similarly to CME, a weighted sum of canonical features $k_{\mathcal Y}(\cdot,Y_i)$, with weights 

$$
\gamma(t)=\frac{1}{n}(K_{\mathcal T}\circ K_{\mathcal X}+n\lambda I_{n})^{-1}k_{\mathcal T}(\mathbf{T},t)\circ K_{\mathcal X}\mathbf{1}.
$$
We note that the ATE estimator can be obtained by simply using the linear kernel $k_{\mathcal Y}(y,y')=y^\top y'$. When conditioning on the effect modifiers $V$, the relationship between $V$ and other variables needs to be carefully considered to allow for the correct adjustment. The simplest case is when $V$ is a subset of $X$, in which case $\mu_{Y|do(T=t),V=v}=C_{Y|T,X}(k_{\mathcal{T}}(\cdot,t)\otimes\mu_{X|V=v})$.

\subsection{Backdoor weights and Backdoor HSIC}
Recall that we observe data $\{(X_i,T_i,Y_i)\}_{i=1}^n$ from the observational distribution $p(X,T,Y)=p(X)p(T|X)p(Y|X,T)$, but are interested in the interventional distribution $p^{\star}(X,T,Y)=p(X)p^{\star}(T)p(Y|X,T)$. This means we can estimate quantities under $p^\star$ using the density ratios applied to observational data as importance weights,
\begin{equation}
\label{eq:densityratios}
w(T,X)=\frac{p^{\star}(X,T,Y)}{p(X,T,Y)}=\frac{p^{\star}(T)}{p(T|X)}.
\end{equation}
For example, it is easy to see that for any real-valued functions $f$, $g$, 
$\mathbb{E}_{p^\star}\left[f(T)g(Y)\right]=\mathbb{E}_{p}\left[w(T,X)f(T)g(Y)\right]$.
Note that we have modified the marginal distribution of treatment $T$ (from $p(T)$ to $p^\star(T)$), which might be important to control the variance of importance weights. An example of this approach is the Backdoor HSIC \cite{huKernelTestCausal2024}, where the cross-covariance operator $C^\star_{TY}$ between the treatment $T$ and outcome $Y$ under $p^\star$ is estimated with backdoor weights. This is based on the following observation
$$
C^\star_{TY}=\mathbb{E}_p\left[w(T,X)k_{\mathcal T}(\cdot,T)\otimes k_{\mathcal Y}(\cdot,Y)\right]-\mathbb{E}_{p^{\star}}\left[k_{\mathcal T}(\cdot,T)\right]\otimes \mathbb{E}_{p}\left[w(T,X)k_{\mathcal Y}(\cdot,Y)\right].
$$
Backdoor HSIC can now serve as the statistic for a nonparametric causal association test: is there a dependence between $T$ and $Y$ under the interventional regimes $p^\star$? A permutation-based testing approach is developed in \cite{huKernelTestCausal2024}.
Of course, estimating density ratios in \eqref{eq:densityratios} can itself be a challenging problem for general $T$ and $X$. An effective approach to this, as demonstrated in \cite{huKernelTestCausal2024}, is to use a noise contrastive deep learning method \cite{gutmannNoiseContrastiveEstimationUnnormalized2012}. Note that this also gives us an alternative, weighting-based estimator of IMO, via
$\hat C_{Y|do(T)}=\hat C^\star_{YT}(\hat C^\star_{TT}+\lambda I)^{-1}$.

\section{Causal Data Fusion}
\label{sec:fusion}
Many scenarios of causal inference involve multiple stages of estimation where, within a given stage, estimation does not involve all variables present in the DAG. This can be exploited in the context of \textit{causal data fusion} where multiple unmatched datasets may correspond to subgraphs in a given DAG. Uncertainty becomes particularly important in the causal data fusion settings, as data may be collected from multiple studies with varying data quality and quantity. A Bayesian model for kernel embeddings was applied to estimating average treatment effects (ATE) in causal data fusion in \cite{chauBayesIMPUncertaintyQuantification2021}. Importantly, uncertainty propagates into the estimates of ATE and can guide Bayesian optimization -- a technique which allows finding the values of the treatment variable (e.g. medication dosage), which optimize ATE. 

\begin{figure}
\begin{centering}
\includegraphics[width=0.5\columnwidth]{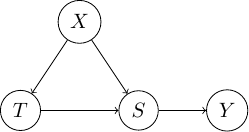}
\par\end{centering}
\caption{\label{fig:bayesimp}A simple graph for the causal data fusion setting. We do not require that the treatment $T$ and outcome $Y$ are matched in our data.}
\end{figure}

We will illustrate this idea on the simple graph given in Fig. \ref{fig:bayesimp} but the ideas can be applied much more broadly.
While \cite{chauBayesIMPUncertaintyQuantification2021} considers the standard ATE $\mathbb E[Y|do(T=t)]$, here we extend the formalism to the estimation of the DTE $\mu_{Y|do(T=t)}$.

Defining $g(y,S)=\mu_{Y|S}(y)=\mathbb{E}\left[k_{\mathcal{Y}}(y,Y)|S\right]$ which is a random variable as a function of $S$, and is an evaluation of the CME at $y\in \mathcal{Y}$, we obtain for the DAG in Fig. \ref{fig:bayesimp}, following similar steps as in \cite{chauBayesIMPUncertaintyQuantification2021},
\begin{align}
\nonumber
\mathbb{E}\left[k_{\mathcal{Y}}(y,Y)|do(T=t)\right]&=\int \mathbb{E}\left[k_{\mathcal{Y}}(y,Y)|S\right]p(S|do(T=t))dS\\\nonumber
&=\int g(y,S)p(S|do(T=t))dS\\\nonumber
&=\int \langle g(y,\cdot),k_{\mathcal S}(\cdot,S) \rangle_{\mathcal{H}_{\mathcal S}} p(S|do(T=t))dS\\\nonumber
&=\langle g(y,\cdot),\int k_{\mathcal S}(\cdot,S)  p(S|do(T=t))dS \rangle_{\mathcal{H}_{\mathcal S}}\\
&=\langle g(y,\cdot),\mu_{S|do(T=t)} \rangle_{\mathcal{H}_{\mathcal S}}.
\label{eq:fusion}
\end{align}
From \eqref{eq:fusion}, it is clear that it suffices to estimate the CME $\mu_{Y|S=s}$ on a dataset $\mathcal{D}_{1}=\{S^{(1)}_{i},Y_{i}\}_{i=1}^{n_{1}}$, and the interventional mean embedding $\mu_{S|do(T=t)}$ (using backdoor adjustment) on a separate dataset $\mathcal{D}_{2}=\{X_{j},T_{j},S_{j}^{(2)}\}_{j=1}^{n_{2}}$. Furthermore, plugging in the backdoor formula, we can express the above in terms of CMOs as
\begin{align*}
\mathbb{E}\left[k_{\mathcal{Y}}(y,Y)|do(T=t)\right]&=\langle g(y,\cdot),\mu_{S|do(T=t)} \rangle_{\mathcal{H}_{\mathcal S}}\\
&=\langle C_{Y|S}^{*}k_{\mathcal{Y}}(\cdot,y),C_{S|T,X}(k_{\mathcal{T}}(\cdot,t)\otimes\mu_{X}) \rangle_{\mathcal{H}_{\mathcal S}}\\
&=\langle k_{\mathcal{Y}}(\cdot,y),C_{Y|S}C_{S|T,X}(k_{\mathcal{T}}(\cdot,t)\otimes\mu_{X}) \rangle_{\mathcal{H}_{\mathcal Y}},
\end{align*}
whereby $\mu_{Y|do(T=t)}=C_{Y|S}C_{S|T,X}(k_{\mathcal{T}}(\cdot,t)\otimes\mu_{X})$.
This indicates that the key objects of interest are again the CMOs and that the DTE in this case can be obtained by simply composing the respective CMOs.

 We note here that the treatment $T$ and outcome $Y$ of interest do not appear in the same observational study and yet our model is able to estimate causal effect of $T$ on $Y$. In fact, such combining of embedding estimators from their corresponding datasets can proceed in a wide range of scenarios, so that each individual estimator entering the inner product in \eqref{eq:fusion} arises from its own potentially complex estimation setting such as those discussed in Sections \ref{sec:frontdoor}-\ref{sec:proxies}.   Namely, the assumption of fully observed confounding discussed so far is unrealistic in many applications. There are a variety of specific settings where, under additional assumptions regarding the causal graphical structure of the problem at hand, it is still possible to estimate treatment effects in the presence of unobserved confounders. Here, we shall discuss three such settings: frontdoor, instruments, and proxies.

\section{Frontdoor adjustment}
\label{sec:frontdoor}
\begin{figure}
\begin{centering}
\includegraphics[width=0.6\columnwidth]{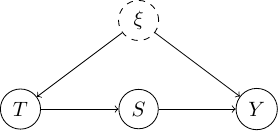}
\par\end{centering}
\caption{\label{fig:frontdoor}A DAG satisfying the frontdoor criterion}
\end{figure}

A frontdoor setting is illustrated in Fig.\ref{fig:frontdoor}, where a set of variables $S$ is the \textit{mediator} of the effect of $T$ on $Y$. This means that all causal paths from $T$ to $Y$ go through $S$. Unobserved confounder $\xi$ directly causes $T$ and $Y$ only. In this case, it is a standard result of do-Calculus that
\[
p(Y|do(T=t))=\int\int p(Y|S,T')p(T')dT'p(S|T=t)dS.
\]

The IME $\mu_{Y|do(T=t)}$ can be expressed in terms of the IME $\mu_{Y|do(S=s)}$, which can be obtained using backdoor since $T$ is the valid backdoor adjustment set for the effect of mediator $S$ on outcome $Y$, and the CME $\mu_{S|T=t}$. Write $g(y,\cdot)=\mu_{Y|do(S=\cdot)}(y)=C^{*}_{Y|do(S)}k_{\mathcal{Y}}(\cdot,y)$, where $C_{Y|do(S)}:\mathcal H_{\mathcal S}\to\mathcal H_{\mathcal Y}$ is the IMO corresponding to backdoor adjustment, i.e. its action is defined as $C_{Y|do(S)}k_{\mathcal{S}}(\cdot,s):=C_{Y|S,T}(k_{\mathcal{S}}(\cdot,s)\otimes\mu_{T})$. We see that
\begin{align*}
\mu_{Y|do(T=t)}(y)&=\langle g(y,\cdot),\mu_{S|T=t} \rangle_{\mathcal{H}_{\mathcal{S}}}\\
&=\langle C^{*}_{Y|do(S)}k_{\mathcal{Y}}(\cdot,y),C_{S|T}k_{\mathcal{T}}(\cdot,t) \rangle_{\mathcal{H}_{\mathcal{S}}}\\
&=\langle  k_{\mathcal{Y}}(\cdot,y),C_{Y|do(S)}C_{S|T}k_{\mathcal{T}}(\cdot,t) \rangle_{\mathcal{H}_{\mathcal{Y}}}, 
\end{align*}
and hence
$$
\mu_{Y|do(T=t)}=C_{Y|do(S)}C_{S|T}k_{\mathcal{T}}(\cdot,t)=C_{Y|do(S)}\mu_{S|T=t}.
$$

Now, recalling that $\mu_{S|T=t}=\int k_{\mathcal{S}}(\cdot,S)p(S|T=t)dS\in \mathcal H_{\mathcal S}$, we see by linearity of CMOs that

\begin{align*}
\mu_{Y|do(T=t)}&=C_{Y|do(S)}\mu_{S|T=t}\\
&=C_{Y|do(S)}\int k_{\mathcal{S}}(\cdot,S)p(S|T=t)dS\\
&=\int (C_{Y|do(S)}k_{\mathcal{S}}(\cdot,S))p(S|T=t)dS\\
&=\int C_{Y|S,T}(k_{\mathcal{S}}(\cdot,S)\otimes\mu_{T})p(S|T=t)dS\\
&=C_{Y|S,T}\left( \int k_{\mathcal{S}}(\cdot,S)p(S|T=t)dS \otimes\mu_{T}\right)\\
&=C_{Y|S,T}(\mu_{S|T=t}\otimes\mu_{T})\\
&=C_{Y|S,T}((C_{S|T}k_{\mathcal{T}}(\cdot,t))\otimes\mu_{T}).
\end{align*}

\begin{figure}[ht]
    \centering
    \begin{subfigure}[b]{0.5\textwidth}
        \centering
        \includegraphics[width=\textwidth]{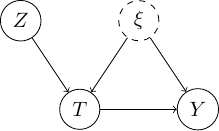}
        \caption{A DAG with a valid instrument $Z$.}
        \label{fig:instrument}
    \end{subfigure}
    \hfill
    \begin{subfigure}[b]{0.45\textwidth}
        \centering
        \includegraphics[width=\textwidth]{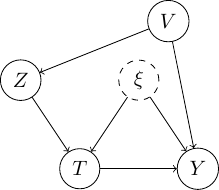}
        \caption{A DAG with a valid conditional instrument $Z$, where $V$ are effect modifiers.}
        \label{fig:cinstrument}
    \end{subfigure}
    
    \caption{Instruments}
    \label{fig:instrument_and_cinstrument}
\end{figure}

\section{Instruments}

\label{sec:instruments}
Instrumental variables are a classical framework to estimate causal effects under unobserved confounding \cite{imbensIdentificationEstimationLocal1994}. A set of variables $Z$ is said to be a valid \textit{instrument} relative to the treatment $T$ and the outcome $Y$, if the following assumptions are satisfied: (a) $Z$ has a causal effect on $T$ (\textit{relevance}), (b) causal effect of $Z$ on $Y$ is fully mediated by $T$ (\textit{exclusion restriction}), (c) there are no backdoor paths from $Z$ to $Y$ (\textit{instrumental unconfoundedness}). This setting is illustrated in Fig. \ref{fig:instrument}. It should be noted that the instrumental variables do not allow us to nonparametrically identify the causal effect, so we must make some additional assumptions. Let us first consider the case of ATE, i.e. estimating $f(t)=\mathbb{E}\left[Y|do(T=t)\right]$. We assume that unobserved confounders enter additively into the outcome, i.e. that
\[
Y=f(T)+\xi.
\]
where $\mathbb{E}\left[\xi\right]=0$, but $\mathbb{E}\left[\xi|T\right]\neq 0$.
Since $Z$ is a valid instrument, we know that $\xi$ and $Z$ are marginally independent so
\begin{equation}
    \mathbb{E}\left[Y|Z\right]=\mathbb{E}\left[f(T)|Z\right]+\underbrace{ \mathbb{E}\left[\xi|Z\right] }_{ =0 }.
\label{eq:instrument}
\end{equation}
The key idea now is to solve \eqref{eq:instrument} for $f$. We note that this is the problem of deconditioning, discussed in Section \ref{sec:deconditioning}. Here we make an explicit connection to DMOs as well as extend the framework to DTE and CDTE. We modify the assumption to
\[
k_{\mathcal{Y}}(y,Y)=f(y,T)+\xi(y),
\]
for some zero-mean stochastic process $\xi$ taking values in $\mathcal{H}_\mathcal{Y}$ (marginally independent of $Z$). Then
\[
\mu_{Y|Z}(y)=\mathbb{E}\left[k_{\mathcal{Y}}(y,Y)|Z\right]=\mathbb{E}\left[f(y,T)|Z\right]+\underbrace{ \mathbb{E}\left[\xi(y)|Z\right] }_{ =0 }.
\]

Hence, we need to find the function $f:\mathcal Y\times \mathcal T \to\mathbb R$ such that $g(y,z)=\mu_{Y|Z=z}(y)=\mathbb{E}\left[f(y,T)|Z=z\right]$, so we can express this using DMO as $D^*_{T|Z}g(y,\cdot)=f(y,\cdot)$. 
But $g(y,\cdot)=C_{Y|Z}^{*}k_{\mathcal{Y}}(\cdot,y)$. This means that we can express the IME as
\begin{align*}
\mu_{Y|do(T=t)}(y)&=f(y,t)\\
&=\langle k_{\mathcal T}(\cdot,t), f(y,\cdot) \rangle_{\mathcal{H}_{\mathcal{T}}}\\
&=\langle k_{\mathcal T}(\cdot,t), D^{*}_{T|Z}C^{*}_{Y|Z}k_{\mathcal{Y}}(\cdot,y)\rangle_{\mathcal{H}_{\mathcal{T}}}\\
&=\langle C_{Y|Z}D_{T|Z}k_{\mathcal T}(\cdot,t),k_{\mathcal{Y}}(\cdot,y) \rangle_{\mathcal{H}_{\mathcal{Y}}}. 
\end{align*}

We note that the instrumental variables can be readily extended (and are often applied) to the case of CATE estimation by simply allowing that $f$ also depends on $V$, where $V$ are observed variables assumed to block any backdoor paths between $T$ and $V$ (Fig. \ref{fig:cinstrument}). In this case, $Z$ is known as the conditional instrument. Analogous derivations to above give 
$\mu_{Y|do(T=t),V=v}(y)=C_{Y|Z}D_{(T,V)|Z}(k_{\mathcal T}(\cdot,t)\otimes k_{\mathcal V}(\cdot,v))$. CATE estimation with instruments and kernel embeddings was also studied in \cite{xuLearningDeepFeatures2021}. While they do not use the tools of deconditioning that we adopt here, they recover similar estimators via the following two-stage approach: (1) estimate CME for $(T,V)|Z$, and (2) find $f$ by minimizing
\[
\min_{f}\mathbb{E}_{Y,Z}\left[(Y-\mathbb{E}_{T,V|Z}\left[f(T,V)\right])^{2}\right],
\]
where the inner expectation makes use of the estimated CME $\mu_{T,V|Z}$.

\section{Proxies}
\label{sec:proxies}

\begin{figure}
\begin{centering}
\includegraphics[width=0.5\columnwidth]{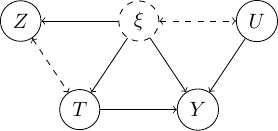}
\par\end{centering}
\caption{\label{fig:proxies}The setting for proximal causal learning. The edges between $Z$ and $T$ and between $\xi$ and $U$ can be in either direction.}
\end{figure}

Proximal method for causal learning is the way to recover causal effect under unobserved confounding under structural assumptions of the DAG given in Fig.\ref{fig:proxies} . It was proposed in \cite{miaoIdentifyingCausalEffects2018} and kernel-based estimators were studied in \cite{mastouriProximalCausalLearning2021} followed by a deep feature version in \cite{xuDeepProxyCausal2021}. For simplicity, we shall first study ATE $\mathbb E[Y|do(T=t)]$ and assume that $\mathcal Y\subseteq \mathbb R$. As before, the confounder $\xi$ is unobserved but we assume that we do observe a \textit{treatment proxy} $Z$ and an \textit{outcome proxy} $U$, such that:
\begin{equation}
    U\perp\!\!\!\!\perp (Z,T)|\xi,\quad Y\perp\!\!\!\!\perp Z|(T,\xi).
    \label{eq:proxy_assumptions}
\end{equation}
\cite{miaoIdentifyingCausalEffects2018} show that under some reasonable identifiability assumptions, we can still recover ATE. As we will see, the proximal setting at its core also relies on deconditioning. Since $\xi$ is a sufficient adjustment set, from backdoor adjustment, we have
\begin{equation}
\mathbb E(Y|do(T=t))=\int \mathbb E(Y|T=t,\xi)p(\xi)d\xi,
\label{eq:backdoor_xi}   
\end{equation}
but of course $\xi$ is unobserved. However, assume we can solve for some function $h:\mathcal T \times \mathcal U \to \mathbb R$ such that 
\begin{equation}
\mathbb{E}(Y|T,Z)=\int h(T,U)p(U|T,Z)dU.
\label{eq:fredholm_ate}
\end{equation}
We refer to $h$ as the \textit{bridge function} and the equation \eqref{eq:fredholm_ate} as the \textit{proximal Fredholm equation}. Structural assumptions \eqref{eq:proxy_assumptions} imply
\[
p(U|T,Z)=\int p(U|\xi)p(\xi|T,Z)d\xi,\quad \mathbb{E}(Y|T,Z)=\int \mathbb{E}(Y|T,\xi)p(\xi|T,Z)d\xi.
\]
Putting these together gives
\[
\int \mathbb{E}(Y|T,\xi)p(\xi|T,Z)d\xi=\int\left( \int h(T,U) p(U|\xi)dU \right)p(\xi|T,Z)d\xi.
\]
Under certain assumptions (cf. \cite{miaoIdentifyingCausalEffects2018}), we can equate the integrands to obtain
\[
\mathbb{E}(Y|T,\xi)=\int h(T,U) p(U|\xi)dU,
\]
and plugging into \eqref{eq:backdoor_xi} gives the ATE
\begin{align*}
\mathbb{E}(Y|do(T=t))&=\int\int h(t,U) p(U|\xi)p(\xi)dUd\xi
\\&=\int h(t,U) p(U)\underbrace{ \left( \int p(\xi|U)d\xi \right) }_{ =1 }dU
\\&=\int h(t,U)p(U)dU.
\end{align*}
Hence, solving for the bridge function $h$ in \eqref{eq:fredholm_ate}, which is a deconditioning task, is the key part of ATE estimation in this setting.

Turning our attention to DTE, define $g(y,t,z)=\mu_{Y|T=t,Z=z}(y)$, so that $g(y,\cdot)=\mathbb{E}\left[k_{\mathcal{Y}}(Y,y)|(T,Z)=\cdot\right]=C_{Y|T,Z}^{*}k_{\mathcal{Y}}(\cdot,y)$ and the corresponding proximal Fredholm equation is 
\begin{equation}
g(Y,T,Z)=\int h(Y,T,U)p(U|T,Z)dU.
\end{equation}
This means that $h$ is the result of deconditioning, so $D_{(T,U)|(T,Z)}^*g(y,\cdot)=h(y,\cdot)$, and hence the bridge function is
$$h(y,t,u)=\langle k_{\mathcal T}(\cdot,t)\otimes k_{\mathcal U}(\cdot,u), D^{*}_{(T,U)|(T,Z)}C^{*}_{Y|T,Z}k_{\mathcal{Y}}(\cdot,y)\rangle_{{\mathcal H}_{\mathcal T\times \mathcal U}}.$$
Finally, $U$ can be integrated out through the marginal $p(U)$, giving the IME
\begin{align*}
\mu_{Y|do(T=t)}(y)&=\int h(y,t,U)p(U)dU\\
&=\int\langle k_{\mathcal T}(\cdot,t)\otimes k_{\mathcal U}(\cdot,U), D^{*}_{(T,U)|(T,Z)}C^{*}_{Y|T,Z}k_{\mathcal{Y}}(\cdot,y)\rangle p(U)dU\\
&= \langle k_{\mathcal T}(\cdot,t)\otimes \mu_{U}, D^{*}_{(T,U)|(T,Z)}C^{*}_{Y|T,Z}k_{\mathcal{Y}}(\cdot,y)\rangle_{\mathcal H_{\mathcal T\times \mathcal U}}\\
&=\langle C_{Y|T,Z}D_{(T,U)|(T,Z)}(k_{\mathcal T}(\cdot,t)\otimes \mu_{U}),k_{\mathcal{Y}}(\cdot,y) \rangle _{\mathcal H_{\mathcal Y}}.
\end{align*}

\section{Discussion}

We reviewed how kernel embeddings can be used in the context of estimation of average treatment effects and distributional treatment effects in a variety of causal settings. Expressing the distributional treatment effects using conditional and deconditional mean operators allows for a unified framework of constructing nonparametric estimators and studying their properties.

While computational complexity is often cited as a bottleneck of kernel methods, due to the need to invert $n\times n$ kernel matrices, which also features prominently when dealing with conditional mean operators, this concern has been superseded by the rise of large-scale kernel approximation tools, such as random Fourier features \cite{rahimiRandomFeaturesLargescale2007}. Theoretical properties of random Fourier features are now well understood \cite{liUnifiedAnalysisRandom2021}, offering a guaranteed reduction in computational complexity without sacrificing convergence properties and expressivity of the resulting estimators.

Finally, standard kernels (like Gaussian and Mat\'ern families) correspond to fixed pre-specified representations of data in a given RKHS. This representation has limited expressive capacity when data are complex (text or images), and working with learned representations is preferred. The framework of learned representations, however, is fully compatible with kernel embeddings and learned deep feature approaches have been explored in a number of papers, e.g. \cite{xuLearningDeepFeatures2021,xuNeuralMeanEmbedding2022}. Namely, the framework of kernel embeddings is agnostic to the representation of the individual data points -- even if those representations are learned and parameterised with deep neural networks, the mathematical framework of embeddings and how to combine them in order to perform causal inference remains the same. 

{\footnotesize
\bibliography{tes_bib}
}

\end{document}